\documentclass[letterpaper, 10 pt, conference]{ieeeconf}  % Comment this line out if you need a4paper

\IEEEoverridecommandlockouts                              % This command is only needed if 
\usepackage{graphicx} % for pdf, bitmapped graphics files
\setkeys{Gin}{width=\linewidth,keepaspectratio}
\usepackage{amsmath} % assumes amsmath package installed
\usepackage{amssymb}
\usepackage{xcolor}
\usepackage{tikz}
\usepackage{float}
\usetikzlibrary{positioning,arrows.meta}
\usepackage{subcaption}
\usepackage{pgfplots}
\pgfplotsset{compat=1.18}
\usepgfplotslibrary{statistics,groupplots}
\usepackage{xcolor}
\usepackage{cite}
\usepackage{bm}
\usepackage{url}

\usepackage{eso-pic}
\AddToShipoutPictureBG*{%
  \AtPageUpperLeft{%
    \put(\LenToUnit{0.5\paperwidth},\LenToUnit{-0.35in}){%
      \makebox[0pt][c]{\parbox[t]{0.86\paperwidth}{\centering\scriptsize
        This is the accepted version of a paper published in the 2026 IEEE 22nd International
        Conference on Automation Science and Engineering (CASE), Shenyang, China,
        August 17--21, 2026, pp.~966--973. DOI: \url{https://doi.org/10.1109/CASE69030.2026.11704152}\\[2pt]
        \copyright~2026 IEEE. Personal use of this material is permitted. Permission from IEEE
        must be obtained for all other uses, in any current or future media, including
        reprinting/republishing this material for advertising or promotional purposes, creating
        new collective works, for resale or redistribution to servers or lists, or reuse of any
        copyrighted component of this work in other works.}}}}}

\title{\LARGE \bf
RAPID: Robotic Agentic Platform for Intelligent Disassembly --- Platform Design and Unscrewing Evaluation
}

\author{Zachary Allen$^{1,*}$, Max Conway$^{1,*}$, Lyle Antieau${^1}$, Allen Devaraj${^1}$, and Nikolaus Correll$^{1,2}$\thanks{$^{1}$Department of Computer Science, University of Colorado, Boulder, $^{2}$Department of Aerospace and Mechanical Engineering, University of Notre Dame. The * indicates equal contribution.}}

\begin{document}

\maketitle
\thispagestyle{empty}
\pagestyle{empty}

%%%%%%%%%%%%%%%%%%%%%%%%%%%%%%%%%%%%%%%%%%%%%%%%%%%%%%%%%%%%%%%%%%%%%%%%%%%%%%%%
\begin{abstract}
Electric vehicles (EV) create an urgent need for scalable battery recycling, yet disassembly of EV battery packs remains largely manual due to high design variability. We present our Robotic Agentic Platform for Intelligent Disassembly (RAPID), designed to investigate perception-driven manipulation, flexible automation, and AI-assisted robot programming in realistic recycling scenarios. The system integrates a gantry-mounted industrial manipulator, RGB-D perception, and an automated nut-running tool for fastener removal, evaluated on the top-cover unscrewing task. An open-vocabulary object detection pipeline achieves 0.9757~mAP50, enabling reliable identification of screws, nuts, BusBars, and other components. We experimentally evaluate (n=204) three one-shot fastener removal strategies: taught-in poses (97\% success rate, 24min duration), one-shot vision execution (57\%, 29min), and visual servoing (83\%, 36min), comparing success rate and disassembly time for the battery's top cover fasteners. To support flexible interaction, we introduce agentic AI specifications for robotic disassembly tasks, allowing LLM agents to translate high-level instructions into robot actions through structured tool interfaces and ROS services. We evaluate SmolAgents with GPT-4o-mini and Qwen 3.5 9B/4B on edge hardware. Tool-based interfaces achieve 100\% task completion, while automatic ROS service discovery shows 43.3\% failure rates, highlighting the need for structured robot APIs for reliable LLM-driven control.  This open-source platform enables systematic investigation of human–robot collaboration, agentic robot programming, and increasingly autonomous disassembly workflows, providing a practical foundation for research toward scalable robotic battery recycling.
\end{abstract}

%%%%%%%%%%%%%%%%%%%%%%%%%%%%%%%%%%%%%%%%%%%%%%%%%%%%%%%%%%%%%%%%%%%%%%%%%%%%%%%%
\section{INTRODUCTION}
The growing adoption of electric vehicles (EVs) poses a significant challenge for recycling their batteries \cite{beaudet2020key}, which contain high-value materials that can often be recovered at rates exceeding 90\% \cite{ali2025sustainable}. Modern EV batteries can reach voltages up to 800V, creating serious safety hazards such as electric shock and difficult-to-extinguish thermal runaways. Manual disassembly is therefore laborious and dull, dirty, and dangerous, making it a prime target for automation. At the same time, economic pressures are rising. Conventional batteries use nickel, manganese, and cobalt (NMC), but newer mass-market lithium iron-phosphate (LFP) batteries have raw material costs less than a quarter of NMC, increasing the need for cost-efficient disassembly solutions.

\begin{figure}[t]
    \centering
    \includegraphics[width=0.98\columnwidth]{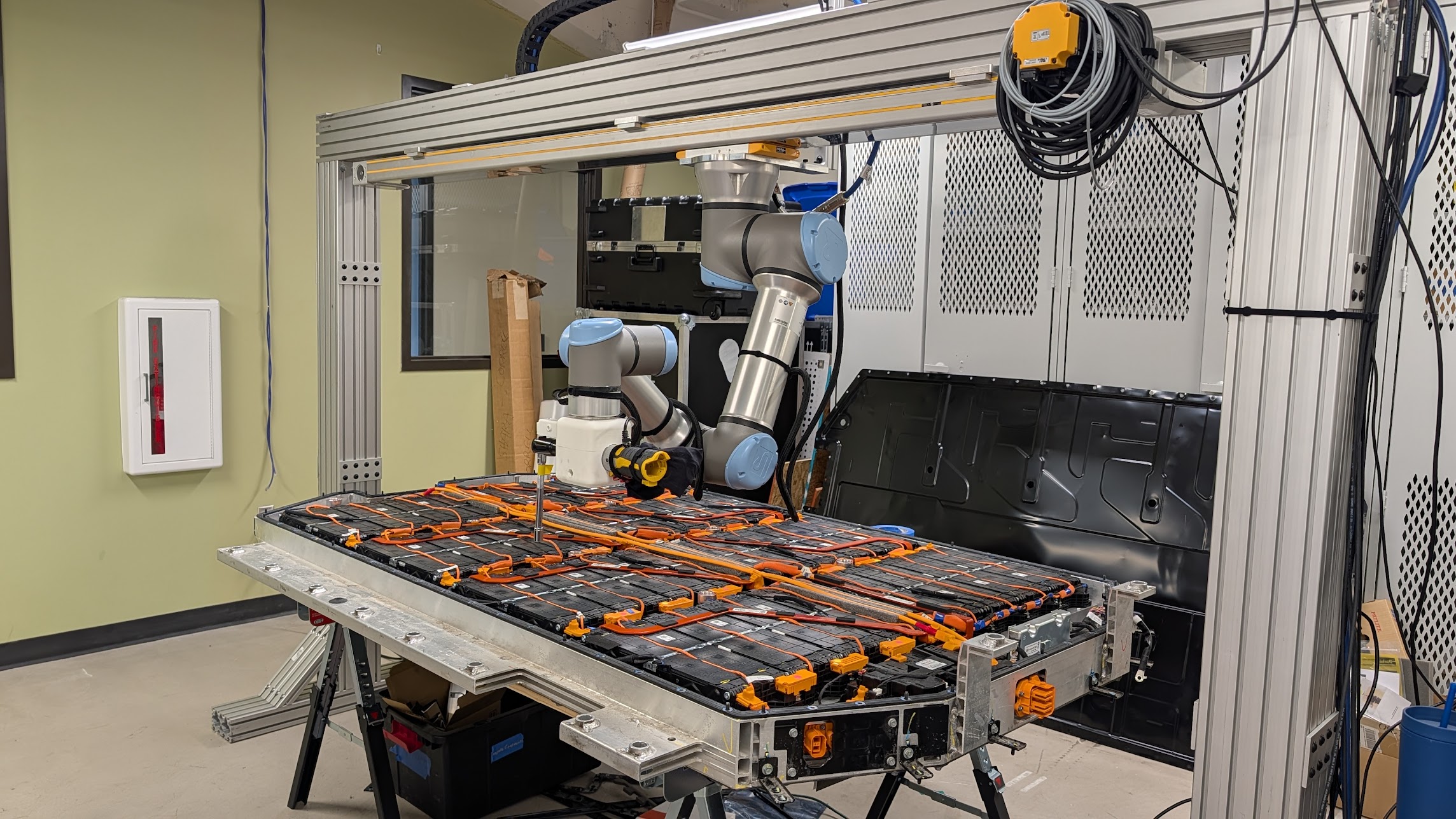}
    \caption{Hyundai Ioniq5 EV battery being dismantled by the RAPID system%using a custom nut-runner mounted on a UR16e collaborative robot and a 2.1m Parker linear gantry.
    \label{fig:real}}
     \vspace{-12pt}
\end{figure}

Shredding batteries into a ``black mass'' for metallurgical recovery is a common recycling approach \cite{CircularMetallurgy}. However, shredding entire batteries,including aluminum frames, copper BusBars, and steel screws lowers economic value \cite{choux2024shred}, and accelerates wear on shredding equipment. These limitations have increased interest in autonomous disassembly techniques. %Emerging techniques, such as laser cutting \cite{rettenmeier2024laser}, are particularly useful for cell-to-pack architectures where cells are glued into large monolithic structures, or when BusBars are welded instead of mechanically connected. Such cutting methods complement robotic disassembly, especially when component reuse is desired.

Fully autonomous battery disassembly requires complex robotic systems capable of a wide range of operations, including screwing, cutting, and prying. Existing approaches for hybrid batteries employ multiple robots with specialized tools \cite{al2024automated,huang2026robotic,tan2025robotic,qu2024robotic}.%, often achieving reduced man-hours despite slower operation times compared to human workers. 
To reduce system complexity, human-robot collaborative approaches where robots handle repetitive screwing while humans manage dexterous tasks have been proposed \cite{chen2014robot,wegener2015robot}.

Given the diversity of EV battery designs and disassembly tasks, we advocate a human-robot collaborative approach that incrementally augments human disassembly stations, combining automation for repetitive operations with flexible human oversight. Here, we are particularly interested in solutions that can easily deal with different battery geometries and visual appearances, and that can be configured with very little specialized training. 

\subsection{Contributions}
This paper presents the following contributions:
\begin{itemize}
    \item A human-robot collaborative research platform (RAPID) for \textit{full-size} EV battery disassembly. We quantitatively evaluate the top-cover unscrewing task; full disassembly automation remains future work.
    \item An open-world perception pipeline to identify screws, nuts, BusBars, and other battery components.
    \item Evaluation of multiple fastener removal strategies, including taught-in poses, vision-guided execution, and visual servoing.
    \item Introduction of agentic AI specifications that allow large language models (LLMs) to translate high-level instructions into structured robot actions, supporting interactive human collaboration and local deployment on edge hardware.
\end{itemize}

\subsection{Related work}
The majority of work on battery recycling is on the much smaller category of hybrid battery packs, typically small enough to fit in the area underneath the trunk, \cite{al2024automated,chen2014robot,wegener2015robot,qu2024robotic}, not on the emerging class of full-size EV battery packs that often span the entire chassis of the car, and operate at lower voltages of 48V compared to the 400-800V range in EV-LIBs. These larger batteries exceed the workspace even of large industrial arms. The disassembly problem of \textit{packs} that consists of multiple \textit{modules}, should also not be confused with the disassembly of individual modules into \textit{cells} \cite{kay2022robotic}. 

In addition to performing specific tasks of the disassembly process \cite{huang2026robotic,tan2025robotic}, robotic disassembly requires task and motion planning. \cite{choux2021task} investigates a task planning framework for the complete disassembly of an Audi e-tron hybrid battery pack. Here, the planner is hard coding disassembly logic such as removing all visible screws before attempting the next step. In \cite{lee2022robot}, a hierarchical planner is employed to disassemble a wooden toybox and an electronic hard-drive with emphasis on human-robot collaboration. Large Language Models (LLM) can also be used to interactively create a plan together with a human user. Here, LLMs can leverage common-sense knowledge about disassembly and also directly issue high-level commands to a robot's sensors and actuators \cite{raptis2025agentic}. In particular, LLMs can be used interactively to let a user refine a robotic disassembly plan \cite{lynch2023interactive}. In \cite{rachwal2025rai}, an Agentic AI framework has been presented that facilitates interactions between an LLM and sensors and actuators controlled by the Robot Operating System (ROS2). In addition to using the LLM to generate high-level plans, they also show using the LLM to detect anomalies during the execution of classical Finite State Machines (FSM), thereby potentially making robots more robust. Finally, \cite{hathaway2024technoeconomic} presents a techno-economic analysis of the battery disassembly process, computing cost savings larger than 40\% when using one Universal Robot arm to support dismantling of the Mitsubishi Outlander hybrid battery.  

\section{SYSTEM OVERVIEW}
\begin{figure}[thpb]
    \centering
    \includegraphics[width=\columnwidth]{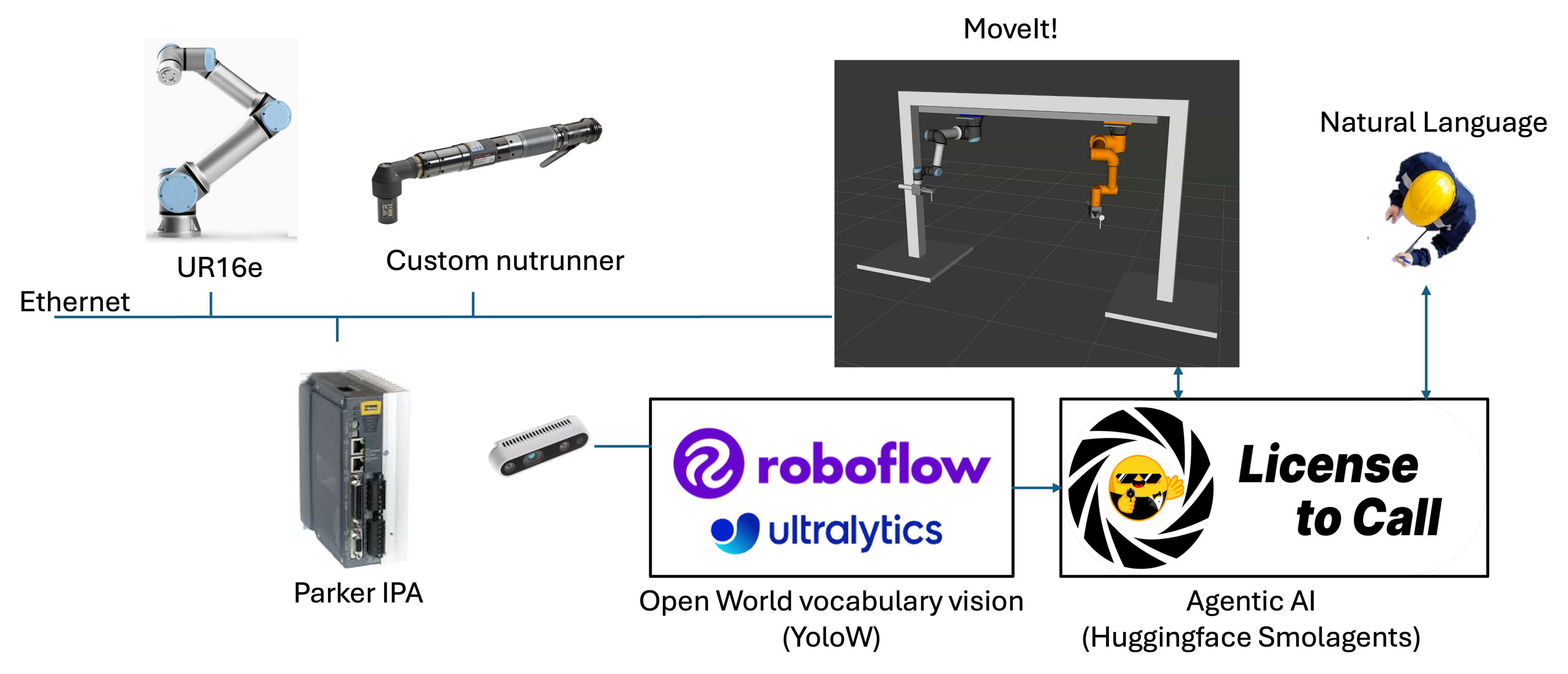}
    \caption{Integration of conventional automation systems, open-world vision systems, motion planning and Agentic AI. \label{fig:overview}}
     \vspace{-12pt}
\end{figure}

 The RAPID system consists of a Universal Robot UR16e, a customized nut-runner (AlloyPower ARW801 with custom Ethernet interface), and a Parker linear gantry.  Figure \ref{fig:real} shows the lab prototype with a Hyundai Ioniq5 full-size EV battery.  The end-effector is equipped with an Intel RealSense D435i. Images are processed via YOLO-World\cite{yoloworld} and motion planning is performed using MoveIt! 2 \cite{coleman2014reducing}. The overall operation is orchestrated using an agentic AI framework SmolAgents \cite{smolagents}. Figure \ref{fig:overview} shows an overview of the system. 

\subsection{Electric Vehicle Battery: Hyundai Ioniq5}
The Hyundai Ioniq5 is a full-size NMC EV battery that is 2.12m in length, 1.22m wide, weighs approximately 450kg, has a capacity of 77.4kWh, and operates at 800V. We have chosen this battery as it can be reversibly assembled and disassembled. Compared with hybrid or plug-in batteries, this kind of battery requires a much larger workspace and additional safety precautions due to the high voltage, but has significantly higher techno-economic benefits due to the increased amount of high-value materials. 

We have performed a complete manual disassembly of the battery to obtain an estimate of manual disassembly time using two untrained workers and a third person for time keeping, who assisted in some of the tasks. The process took 62 minutes. As every task could be done by a single person, we used this data to compute the time required by a single person in \emph{man-minutes}. The battery pack disassembly required a total of 130 man-minutes of labor. Fastener-removal operations (screws, nuts, and bus-bar connections) accounted for 98 man-minutes, representing 75.4\% of the total labor effort. In contrast, all remaining activities—including cover removal, wire harness disconnection, and extraction of the 32 battery modules—required only 32 man-minutes, corresponding to 24.6\% of total labor.
 Times were recorded after performing one complete disassembly and reassembly with the same team, somewhat reducing the effect of learning. Prior to disassembly, we have discharged the battery using a Webasto/AeroVironment AV-900 Heavy Duty Dual Channel Cycling battery cycler to 0V.

\subsection{Robotic System}
\begin{figure*}[t]
\begin{subfigure}{0.75\columnwidth}
    \centering
    \includegraphics[width=\linewidth]{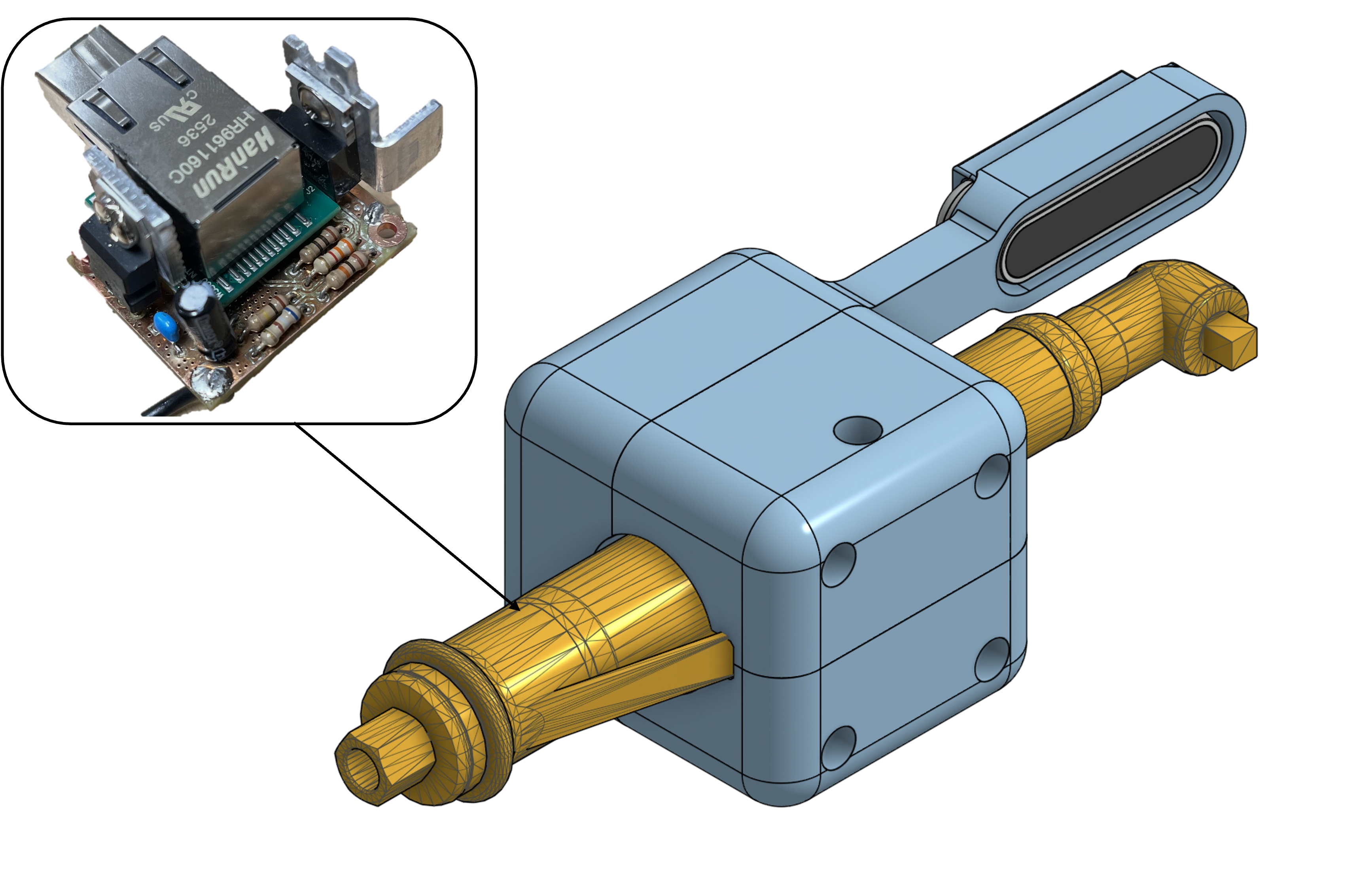}  
\end{subfigure}
\hfill
\begin{subfigure}{1.1\columnwidth}
    \centering
    \resizebox{\linewidth}{!}{
        \begin{tikzpicture}[
  font=\footnotesize,
  node distance=7mm and 10mm,
  block/.style={
    draw, rounded corners,
    align=center,
    text width=2.6cm,
    minimum height=0.8cm
  },
  small/.style={
    draw, rounded corners,
    align=center,
    text width=2.2cm,
    minimum height=0.72cm
  },
  iso/.style={
    draw, rounded corners, dashed,
    align=center,
    text width=2.2cm,
    minimum height=0.72cm
  },
  line/.style={-Latex, thick}
]

% =========================
% Top row (matches your "classic" orientation)
% =========================
\node[block] (pc) {Robot PC\\(ROS2 + agent)};

\node[block, below=3mm of pc, xshift = 6.4cm] (mcu) {Seeed XIAO RP2040 + W5500 Ethernet Module \\Tool interface};

% =========================
% Middle control chain (under MCU, then to tool on right)
% =========================

\node[block, below=5mm of mcu] (drv) {NPN-Driven Mosfet Motor Driver (PWM)\\(BC550B + CSD19534KCS)};
\node[block, right=10mm of mcu] (curr) {W1600 \\Hall Effect \\Current Sensor};
\node[block, right=10mm of drv] (tool) {Nutrunner\\(motor + gearbox)};

% =========================
% Bottom power row (classic left-to-right)
% =========================
\node[small, left=5mm of drv] (dcdc) {3.3V LDO (LD33V)\\(12V $\rightarrow$ 3.3V)};
\node[block, left=5mm of dcdc] (bat) {12V Tool Battery};

% =========================
% Connections
% =========================
% Top row
\draw[dashed] (pc.east) --++ (5cm, 0) -- (mcu.north);

% Drop to isolation and driver
\draw[dashed] (mcu.south) -- (drv.north);

% Driver to tool
\draw[line] (drv.east) -- (tool.west);

% Power chain (battery -> DC/DC -> driver logic)
\draw[line] (bat) -- (dcdc);
\draw[line] (dcdc) -- (drv);
\draw[line] (bat.south) --++ (0,-8mm) --++(6.3cm,0) -- (drv.south);
\draw[line] (dcdc) --++ (0, 1.9cm) -- (mcu);
\draw[dashed] (curr.west)  -- (mcu);
\node[draw,circle,minimum size=6pt,inner sep=0pt,fill=white] (probe) at (8.3,-3.34) {};
\draw[dashed] (probe) --++ (0, 9mm) --++ (19.5mm, 0) -- (curr.south);

% =========================
% Labels (kept off arrows)
% =========================
\node[font=\scriptsize, above=3mm of pc, xshift = 0cm] {ROBOT / CONTROL DOMAIN};
\node[font=\scriptsize, above=0mm of pc, xshift = 0cm] {(earth-referenced)};

\node[font=\scriptsize, above=3mm of bat, xshift = 0mm] {TOOL / BATTERY DOMAIN};
\node[font=\scriptsize, above=0mm of bat, xshift = 0mm] {(floating with battery)};
\node[font=\scriptsize, above=0.5mm of drv, xshift = -11mm] {enable + PWM};
\node[font=\scriptsize, below= 2mm of dcdc] {12V motor power};
\node[font=\scriptsize, left= 4mm of mcu, yshift = 2mm] {3.3V rail};
\node[font=\scriptsize, right= 0mm of mcu, yshift = 4.5mm] {analog};
\node[font=\scriptsize, right= 0mm of mcu, yshift = 2mm] {signal};
\node[font=\scriptsize, right= 10mm of pc, yshift = 2mm] {ethernet (TCP/UDP)};

\end{tikzpicture}
    }
\end{subfigure}
\caption{Left: Custom end-effector for the UR16e that holds a nut-runner and an Intel RealSense D435i in place.    \label{fig:nutrunner} The nut-runner has been retrofitted with an Arduino-based interface board to control speed and measure current via Ethernet. Right: Block diagram of the Ethernet-based nutrunner control system.}
\label{fig:nutrunner_system}
 \vspace{-12pt}
\end{figure*}

RAPID has been dimensioned to service full-size EV batteries and consists of a T-slot arch spanning 2.5m in width, a Parker-Hannifin linear axis (2.1m) and a collaborative Universal Robots 16e 6-DoF robotic arm with a payload of 16 kg.  Both the  Intel RealSense D435i and the nut-runner are mounted using a custom, 3D-printed assembly shown in Figure \ref{fig:nutrunner}. Here, the camera is mounted slightly offset so that the nut-runner does not interfere with its field of view. The nut-runner is mounted so that its torque axis is perpendicular to the last axis of the robot, preventing reaction forces from being absorbed by the wrist.

%\begin{figure}
%    \centering
%    \includegraphics[width=0.7\columnwidth]{nutrunnerassembly.png}
%    \caption{
%}
%\end{figure}

The UR16e provides a socket interface to send URScript commands. We have created custom scripts that allow us to move the robot toward a screw location in force control mode. Once the robot makes contact, the nut-runner is activated. As the nut-runner removes the screw, the robot automatically moves upward as it maintains contact via a constant downward force. 

The nut-runner is equipped with a button that allows for regulating the nut-runner speed by modulating a pulse-width modulated (PWM) signal. We have replicated the built-in circuitry in order to control the nut-runner via an Ethernet connection. The control architecture is shown in Figure \ref{fig:nutrunner_system}. 

The gantry system can be controlled by sending commands via another socket connection. It is equipped with an absolute encoder, which allows precise control with millimeter accuracy. The overall system is controlled by MoveIt! \cite{coleman2014reducing}, which implements path planning and collision avoidance. %To this end, we initially scan the entire battery using the Intel Realsense to create a point-cloud collision object. 

\section{AUTONOMOUS DISASSEMBLY}
\begin{figure}
\includegraphics[width=0.9\columnwidth]{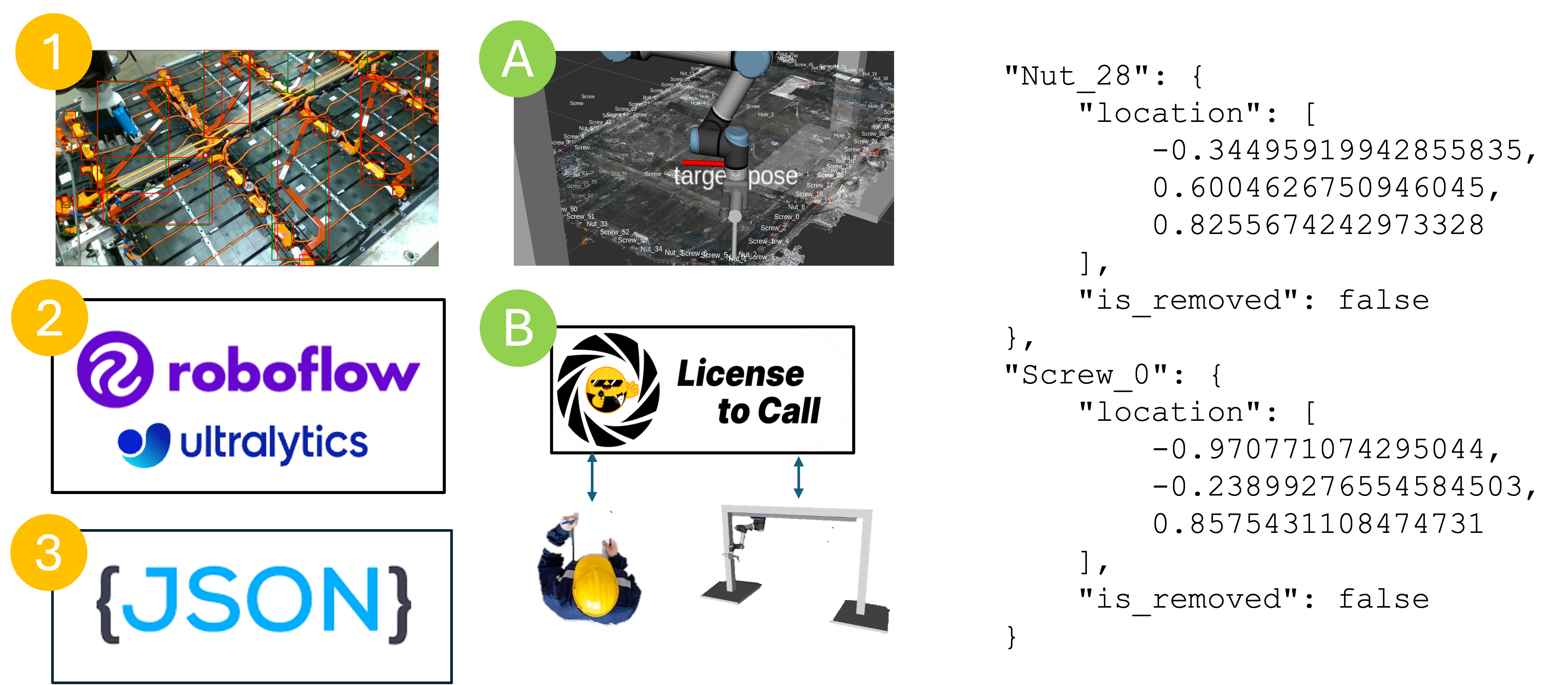}
\caption{Industrial workflow for setup (orange) and execution (green). Left: Relevant battery parts are labeled with text (1) and trained (2). All part locations and disassembly sequence are stored in JSON description (3) and right. Middle: Execution starts with a full scan, registration of parts (A), and uses agentic AI to execute the plan together with a human co-worker (B). Right: Sample JSON data structure.\label{fig:workflow}}
 \vspace{-12pt}
\end{figure}

We envision the following industrial workflow consisting of a setup and an execution phase. The workflow is illustrated in Figure \ref{fig:workflow}.
%
%\begin{enumerate}
%\item Scanning the battery, 
%\item Manually labeling all relevant parts,
%\item Training a vision model,
%\item Map all relevant parts using vision and store their location in a %database,
%\item Manually annotate missing parts,
%\item Create disassembly plans with natural language by providing text input,
%\end{enumerate}
%
%Followed by the actual disassembly sequence,
%\begin{enumerate}
%\item Scan the entire battery and create a collision object,
%\item Use vision to register the battery pose,
%\item Execute the disassembly plan by issuing commands to both the robot and the worker,
%\item Wait for worker input to continue the disassembly plan.
%\end{enumerate}

The resulting data structure contains YOLO-World labels and uses nesting to indicate tasks in the group $T_s = \{t_{s,1}, \dots, t_{s,n}\}$, the set of disassembly sub-tasks associated with step $s \in \mathcal{S}$, a set containing all consecutive disassembly steps. Unscrewing operations dominate the disassembly process, accounting for 10 of the 12 steps; the top-cover unscrewing step alone contains 70 sub-tasks. Figure \ref{fig:workflow} shows a JSON representation of two of these. We assume that all tasks within a group $T_s$ can be performed interchangeably (no precedence constraints), whereas the order of elements in $\mathcal{S}$ governs the overall disassembly sequence. 

Once a model to recognize individual battery elements has been trained, it can be used to identify part locations as shown in Figure \ref{fig:overview}A. Here, false positives need to be removed interactively by the user, whereas false negatives will need to be added by teaching their positions manually by moving the arm to the appropriate location. A user study on creating disassembly plans using natural language is beyond the scope of this paper, and we assume the disassembly plan to be complete for the remainder of this paper.

\subsection{Planning}\label{sec:planning}
We formulate the sequencing of disassembly tasks $s \in \mathcal{S}$ as a Traveling Salesman Problem (TSP) defined over task poses in $SE(3)$. Here, task poses are either camera locations for performing a scan or poses where unscrewing operations take place. Let $t_{s,0}$ represent the robot home configuration and
$t_{s,i} = (p_{s,i}, q_{s,i}) \in SE(3)$ 
where $p_{s,i} \in \mathbb{R}^3$ is the Cartesian position and $q_{s,i} \in \mathbb{H}$ is a unit quaternion representing orientation. The augmented node set is $V_s = \{t_{s,0}\} \cup T$.

We construct a weight matrix
$W_{s,ij} \in \mathbb{R}_{\ge 0}, \quad i,j \in V_s$
where each entry represents the Euclidean distance between task poses $t_{s,i}$ and $t_{s,j}$. 
%
%Specifically, let
%$$
%\mathcal{Q}_{s,ij} = \{ q_{s,ij}^{(1)}, \dots, q_{s,ij}^{(N)} \}
%$$
%denote a set of $N$ inverse kinematic solutions for reaching pose $t_{s,j}$ from pose $t_{s,i}$, obtained via RRT* in joint space. The edge weight is then defined as
%$$
%W_{s,ij} = \min_{k \in \{1,\dots,N\}} \; c\!\left(q_{s,ij}^{(k)}\right)
%$$
%where $c(\cdot)$ denotes the joint-space path cost (e.g., path length, time, or energy). 
Note that $W_{s,ij}$ can be computed offline as the set of tasks $T$ is the same for every battery of a given type. The TSP decision variables are $x_{s,ij} \in \{0,1\}, \quad i,j \in V$ with $x_{s,ij}=1$ if the robot moves directly from $t_{s,i}$ to $t_{s,j}$. The optimization problem is
$$
\min_{x_{s,ij}} \sum_{i \in V} \sum_{j \in V} W_{s,ij} \, x_{s,ij},
$$
%subject to the degree constraints
%$$
%\sum_{j \in V_s} x_{s,ij} = 1,  \quad
%\sum_{i \in V_s} x_{s,ij} = 1 \quad \forall i,j \in V,
%$$
and standard subtour elimination constraints ensuring a single Hamiltonian circuit. This formulation decouples global task sequencing from local motion planning and can be solved using Google's OR tools \cite{cuvelier2023or}. We are using this solution to generate optimal orderings for each task sequence $T_s$.

\subsection{Analytical Inverse Kinematics}\label{sec:ik}
Inverse kinematics (IK) and path planning are non-trivial in a gantry-based system, as poses that are close in Cartesian space can be far apart in joint space when the linear axis is involved.  Standard iterative solvers such as KDL, TracIK, and PICK-IK in MoveIt! return a single,
seed-dependent solution per call when applied to a redundant system and fail occasionally. 
For a 6+1-DoF manipulator, this is insufficient: a single call cannot enumerate
the qualitatively different configurations that arise from varying the gantry
position or selecting a different elbow/shoulder family, leading to low
solver success rates.
Extending to multiple random seeds improves coverage but incurs significant
per-call overhead.
We instead implement a closed-form solver that exploits the UR16e's spherical
wrist to decouple position and orientation subproblems and explicitly
enumerates all kinematic families using a multi-objective cost function that encodes joint distance $d_J(\bm{q}_c, \bm{q}_k) = \sqrt{\sum_{i=1}^{7} w_i (q_{c,i} - q_{k,i})^2}$, where $w_1 = 0.1$ (gantry) and $w_{2\text{--}7} = 1.0$ (arm joints), penalizing large joint-space motions while permitting gantry travel, a proximity penalty $p_P = 1/d_{\text{ee}}$, where $d_{\text{ee}}$ is the Euclidean distance between the wrist and gantry plate, and a geometric penalty $p_A = 1/A_\triangle$, where $A_\triangle$ is the area of the triangle formed by the gantry plate, wrist, and forearm links. Small triangle areas correspond to near-singular, collapsed arm configurations and serve as a geometric proxy for manipulability.

We additionally compute the Yoshikawa manipulability index \cite{yoshikawa1985manipulability} $\mu = \sqrt{\det(\bm{J}\bm{J}^T)}$, where $\bm{J} \in \mathbb{R}^{6 \times 7}$ is the geometric Jacobian, for each planned configuration. While $\mu$ is not directly incorporated into the cost function, it is recorded as a diagnostic metric to quantify the dexterity of the selected configuration. We observe empirically that the area penalty $p_A$ correlates with manipulability, as collapsed arm triangles coincide with near-singular Jacobians.

%\subsubsection{Corridor Constraint}
To produce predictable, human-safe motions in the collaborative workspace, we consider a constrained end-effector path planner enforcing a rectangular corridor between the start and goal poses. %The corridor is defined as a 3D box whose long axis is aligned with the start-to-goal direction vector, with length equal to the segment distance plus a padding of $2 \times 0.25$\,m, and a square cross-section of $2 \times 0.5$\,m per side. The box is centered at the midpoint between start and goal and enforced as a MoveIt position constraint on the end-effector link.

\subsection{Vision}\label{sec:vision_dissassembly}
To detect battery components, we finetune YOLO-World\cite{yoloworld} on a manually labeled dataset collected from our robot using the Intel RealSense D435i with arbitrary orientations. Our dataset contains 563 annotated images with class instances: 345 Bolts, 465 BusBars, 630 Interior Screws, 462 Nuts, 1175 Orange Covers, 697 Screws, and 840 Screw Holes. We apply the bounding box from the Yolo model to the RGBD image to get the 3D pointcloud of the detection, which is then transformed into the robot's base frame.

Each 3D detection is stored in a KD-tree to enable efficient retrieval and lookup of detected objects. At every time-step a new set of detections is generated, for each of these detections we compute the intersection over union (IoU) of the new detection with nearby prior detections in our KD-tree, if this IoU is above a user-specified threshold $\tau$ we consider it to be the same object and merge the two point clouds.

Our vision system can be queried at anytime to retrieve a set of objects $\mathcal{O} = \{(L_i, C_i, B_i, P_i)\} \mid i \in [0, N]$ where $L_i$ is the label of the object using text, $C_i \in [x_i, y_i, z_i]$ is the centroid of the object $B_i \in [x_{i,1}, y_{i,1}, z_{i,1}, x_{i,2}, y_{i,2}, z_{i,2}]$ is the bounding box containing the object, and $P_i \in (0,1)$ is the probability that the object is present as produced by YOLO-World\cite{yoloworld}. Here, using text as an identifier, unlike class ids, favors seamless integration with a text-based agentic AI framework. 

\subsection{Fastener Removal}\label{sec:removal_dissassembly}
We implement and evaluate three approaches to fastener removal: (1) taught-in poses, where the robot is manually placed so that the nut-runner attaches to the screw/nut, (2) using locations obtained by vision and stored in memory, and (3) memory-based locations in conjunction with visual servoing. 
%
%\begin{enumerate}
%    \item Taught-in Poses
%    \item Removal from memory
%    \item Removal from memory w/ Visual servoing
%\end{enumerate}
%
To remove screws from the taught-in poses, the operator manually moves the arm to each screw to create a set of task poses in $T_s$. To remove screws from the KD-tree memory, the vision system is queried to obtain the set of centroids for each detection $\mathcal{C} = \{C_i\} \mid C_i \in [x_i, y_i, z_i] \ i \in [0,N]$. We then compute the task pose for each centroid as 
\[
\vec{z} =
\begin{bmatrix}
0 \\
0 \\
-1
\end{bmatrix}
\ \ 
\vec{y} =
\begin{bmatrix}
-x_i \\
-y_i \\
0
\end{bmatrix}
\frac{1}{
\left\|
\begin{bmatrix}
-x_i & -y_i & 0
\end{bmatrix}
\right\|}
\ \
\vec{x} =
\frac{\vec{y} \times \vec{z}}
{\|\vec{y} \times \vec{z}\|}
\]

\[
t_{s,i} =
\begin{bmatrix}
\vec{x} & \vec{y} & \vec{z} & C_i \\
0 & 0 & 0 & 1
\end{bmatrix}
\] \qquad
\[
T_S = \{\, t_{s,i} \mid C_i \in S \,\}
\]
This has the effect of making the end effector contact the screw while pointing towards the center of the robot workspace. 

After solving for an optimal sequence using TSP, for each $t_{s_i} \in T_s$ we use MoveIt to plan 1cm above the task pose. From there, we engage our nut runner and apply 10N of downward force to remove each screw using the UR16e's built-in force controller. To engage with the screw/nut, we are taking advantage of the play in the nut-runner extension assembly, which leads to a basic form of spiraling pattern \cite{watson2020autonomous}. This play is exaggerated with a longer extension and lessened with a shorter extension, impacting the spiral radius.

To test visual servoing for fastener removal, we take an additional alignment step based on the RGB channels of the RealSense. After we move above each task pose, we compute the expected pixel location of the end effector by transforming the end effector origin into the camera frame via $^{camera}H_{ee}$ and projecting it using the camera intrinsics $c_x, c_y, f_x,f_y$. 
$$
EE = ^{camera}H_{ee} \times \begin{bmatrix}
    0  & 0 & 0 & 1
\end{bmatrix}^T
$$
$$
target = \begin{bmatrix}
    f_x\frac{EE_x}{EE_z}+c_x \\
    f_y\frac{EE_y}{EE_z}+c_y 
\end{bmatrix} 
=
\begin{bmatrix}
    u_{ee} \\
    v_{ee}
\end{bmatrix}
$$
From there, we compute the center of the detected object's bounding box also in pixel space $BB$. We can then normalize our pixels via
$$
BB=\begin{bmatrix}
u_{bb} \\
v_{bb}
\end{bmatrix}
$$
$$
s = \begin{bmatrix}
    (u_{bb} - c_x)/f_x \\
    (v_{bb} - c_y)/f_y
\end{bmatrix}
\\
s^* = \begin{bmatrix}
    (u_{ee} - c_x)/f_x \\
    (v_{ee} - c_y)/f_y
\end{bmatrix}
$$
We then compute the error as how far the bounding box is from our end effector in normalized pixel space
$$
e = s - s^*
$$
From the error, we can compute a depth estimate $Z$ as the median of all the points in the depth channel in the bounding box. From there, the translation-only interaction matrix is given by \cite{VisualServo1}:
$$
L = \begin{bmatrix}
    \frac{-1}{Z} & 0 & \frac{s^*_x}{Z} \\
    0 & \frac{-1}{Z} & \frac{s^*_y}{Z}
\end{bmatrix}
$$
Finally, given our gain $\lambda$, we can calculate our control signal as 
$$
v_c = -\lambda L^\dagger e
$$
By iteratively calculating and applying the linear velocity $v_c$, we can align the screw with our end effector before applying our downward force.

\subsection{Agentic AI}\label{sec:agenticAI}
We employ SmolAgents \cite{smolagents} to connect language models with robotic capabilities through structured tool calls. The agent is evaluated with GPT-4o mini as well as locally deployed Qwen 3.5 models (9B and 4B parameters) running on an NVIDIA Jetson Thor platform. Robot functionality is exposed either via a ROS-based Model Context Protocol (MCP) server or through manually defined tools that wrap higher-level robot skills; the complete system prompt and tool configuration are provided in the project repository.

\section{RESULTS}

\subsection{Planning}
The solver is evaluated using 21 candidate gantry positions with offset from the current position in 0.1\,m increments across $\pm1.0$\,m, clamped to the
actuator travel limits and applying 8 random arm seed positions, yielding up to 168 candidates per target pose.
Solutions are filtered by joint limits, a minimum end-effector clearance of
0.65\,m from the gantry plate, and a triangle-area validity check on the
wrist--forearm--gantry triangle to reject near-singular configurations. Results of our implementation including success rate (SR) and planning time compared with naive KDL are shown in Table \ref{tab:ik_comparison}

\begin{table}[!ht]
\begin{center}
\begin{tabular}{cccc}
\hline
IK Method & Candidates & SR (\%)$\uparrow$ & Avg.\ Plan Time (s)$\downarrow$ \\
\hline
KDL, 1 seed            & 1         & 25.7          & 1.47  \\
KDL, 50 random seeds   & $\leq$50  & 73.0          & 10.46 \\
Analytical (ours)      & $\leq$168 & \textbf{99.3} & \textbf{0.31} \\
\hline
\end{tabular}
\end{center}
\caption{\label{tab:ik_comparison} IK solver comparison over 1\,000 random goal poses with a 20\,s planning budget.}
 \vspace{-12pt}
\end{table}
Here, weights for the cost function have been evaluated by an ablation study, resulting in a trade-off between success rate and manipulability that is beyond the scope of this paper.

Unconstrained RRT freely explores the joint space to find any collision-free path.
This is fast (avg.\ 0.51\,s) and nearly always succeeds (99.4\%), but the
resulting trajectories are indirect: the sampler may swing the arm through
unpredictable arcs that, while collision-free with the static environment model,
can bring the end-effector close to battery components on the table or pass
through regions that are difficult for a nearby human worker to anticipate.
 The corridor constraint addresses this by requiring the end-effector to remain
inside a tight box aligned with the start-to-goal direction vector throughout
the motion. Corridor constraints improve path legibility at significant planning-time cost; formal human-factors validation is left to future work.
% This produces geometrically straight, legible paths that are 29\% shorter in
% end-effector travel (1.26\,m vs.\ 1.78\,m) and avoid lateral excursions over
% the work surface.
% The cost is a ${\sim}93{\times}$ increase in average planning time (47.4\,s
% vs.\ 0.51\,s), as the constrained sampler must find a feasible path within a
% narrow spatial tube, and a 6\% drop in success rate (93.4\% vs.\ 99.4\%) as some goals cannot be reached while keeping the end-effector inside the box.
%For disassembly scenarios where human workers share the workspace and
%battery components occupy the table surface, the corridor's predictable,
%straight motions justify this planning overhead.

\subsection{Vision}\label{sec:vision_results}
\begin{figure}[!htb] 
    \centering
    \includegraphics[width=1.0\linewidth]{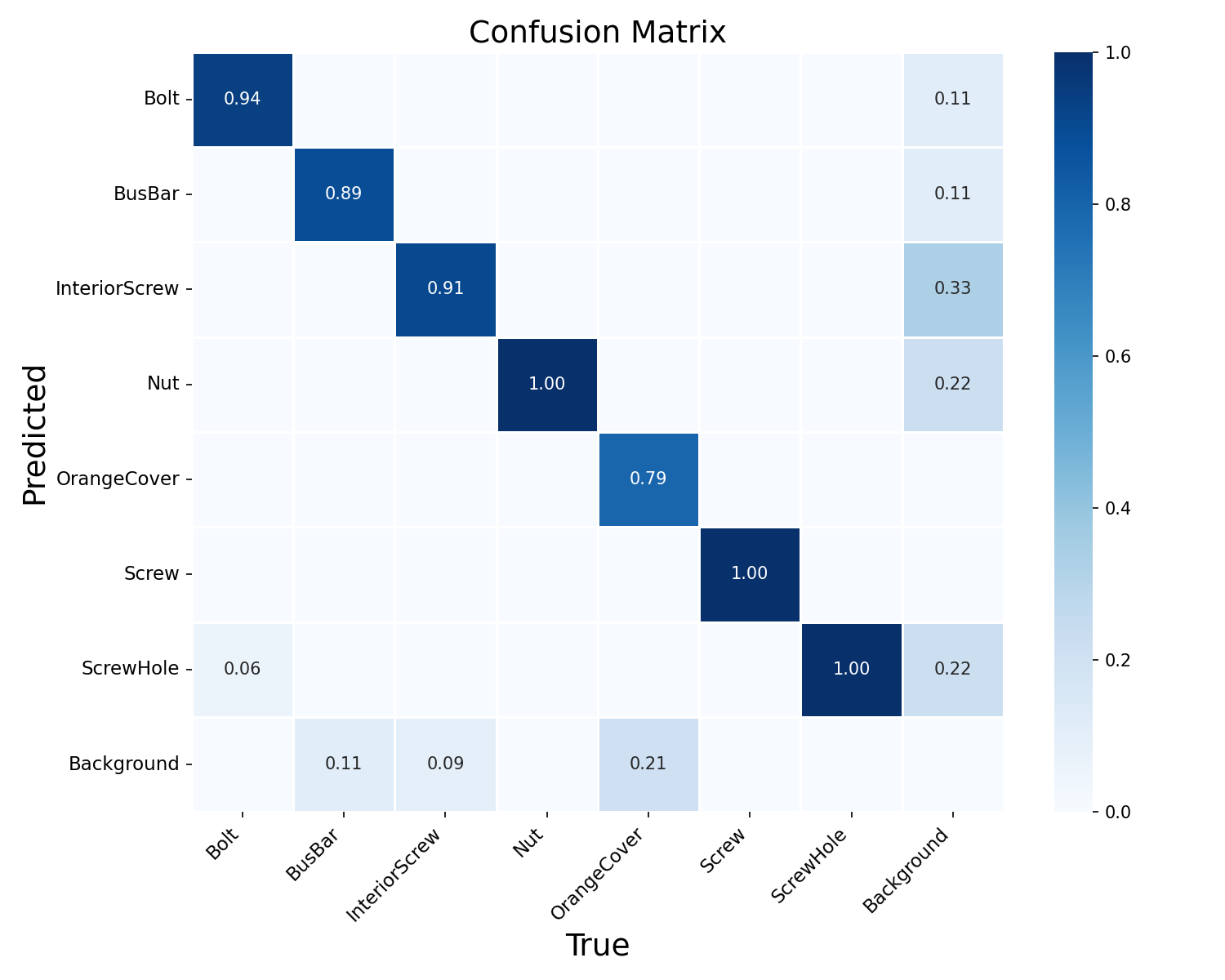}
    \caption{Confusion matrix for YOLO-World on test set containing 18 Bolts, 27 BusBars, 85 InteriorScrews, 40 Nuts, 48 Orange Covers, 69 Screw, and 71 Screw Holes}\label{fig:confusionmatrix}
    \label{fig:placeholder}
     \vspace{-12pt}
\end{figure}

We train and evaluate an ``open world'' vision model YOLO-Worldx\cite{yoloworld} and a conventional class-based model Yolov11L\cite{yolo11_ultralytics} to validate that AgenticAI-friendly open-world models do not compromise accuracy, precision and recall. Results are reported in Table \ref{tab:vision_results}, and show that open-world and class-based systems have comparable and sometimes better performance. The confusion matrix for YOLO-World is shown in Figure \ref{fig:confusionmatrix}. 
\begin{table}[!htb]
\begin{center}
\begin{tabular}{ccc}
\hline
& YOLO-WorldX\cite{yoloworld} & Yolov11L\cite{yolo11_ultralytics} \\
\hline
mAP\@0.5 & \textbf{0.9757} &  0.9708\\
mAP\@50-95 & \textbf{0.7625} & 0.7336\\
Precision & 0.9206 &  \textbf{0.9359}\\
Recall & \textbf{0.9673} & 0.9244\\
\hline
\end{tabular}
\end{center}
\caption{\label{tab:vision_results} Object recognition performance of an LLM-ready open-world ready model (left) vs. class-based vision system (right).}
 \vspace{-12pt}
\end{table}

Our model achieves $~97\%$ recall, requiring users to manually enter only about $~3\%$ of the missed parts. The system produces fewer than $~8\%$ false positives, which must be removed during post-processing. 

%\subsection{System evaluation}
%We have also timed a disassembly sequence for the top cover, by performing disassembly N=\textcolor{red}{3} times removing a total of \textcolor{red}{41*3} screws and \textcolor{red}{27*3} nuts. 

%Additionally we evaluate the systems path planning capabilities and design choices via an ablation study in simulation in which the arm must move between 1000 random pose pairs

\subsection{Fastener Removal}\label{sec:removal_results}
We ran fastener removal for the three configurations: taught-in positions as a baseline, using positions obtained by vision during an initial scan, and using the scanning prior with a visual servoing step.
For teaching a position, we position the nut-runner on the screw/nut. As the assembly contains some play, we validate each position by moving up and running the fastener routine. 

\begin{table}[!ht]
\begin{center}
\resizebox{\columnwidth}{!}{
\begin{tabular}{ccccc}
\hline
& Taught-in Poses & 1-shot vision & Visual Servo  & Human Baseline\\
\hline
Avg Time (min)& \textbf{24.09} & 28.70 & 36.29 & 16.7 \\
Success rate & \textbf{97.06\%} &  57.35\%& 82.84\% & 100\%\\
\hline 
\end{tabular}}
\end{center}
\caption{\label{tab:removal_results} Success rate for different removal strategies (n=204).}
\end{table}

As expected, engaging with the screw based on a pose obtained via vision has a lower success rate than using a taught-in position. While visual servoing can somewhat compensate for this by correcting the pose estimate once the robot is close, it further slows down disassembly. Taught-in poses produce no false positives, whereas the vision-based methods occasionally attempt to remove non-existent fasteners or repeat removals due to duplicate detections. Additionally, the visual servo implementation introduces extra computational overhead from motion planning and repeated inverse kinematics evaluations.

We also evaluate the impact of the length of the nut-runner extension using a short (2.5in) and a long extension (6in) end effector for screw removal, showing a significantly higher success rate with the \emph{longer} extension. The 2.5in extension does not have the length required to generate an effective spiral and/or end-effector compliance, resulting in a significantly worse removal success rate (47.49\%) compared to the 6in extension (97.06\%).

Note that the numbers presented here are single-shot trials, not including retrials, which require accurate detection of failure \cite{watson2023optimal}.

\subsection{Agentic AI}
\begin{figure*}[!htb]
  \centering
  \includegraphics[width=\textwidth]{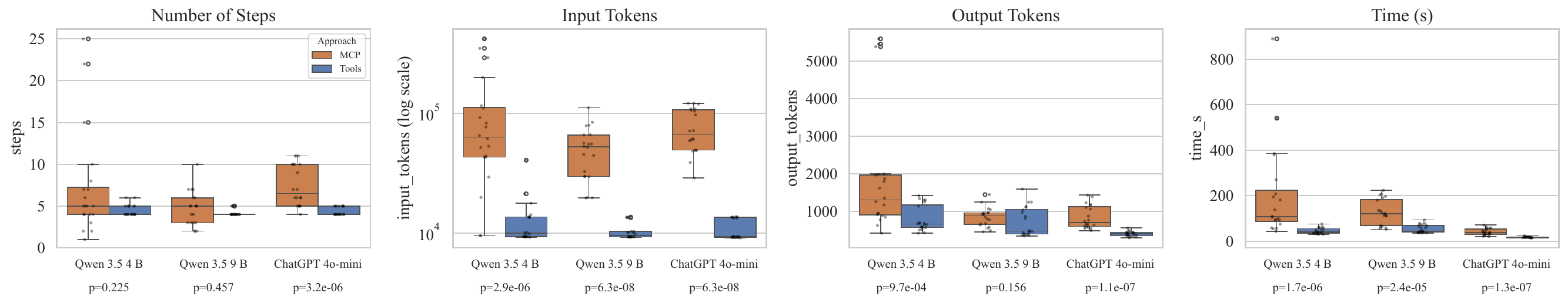}
  \caption{Metrics across three LLM back-ends for agentic execution of
           simple robot-motion commands ``Move the robot 0.2m forward/backward'' via MCP (orange) and an explicit SmolAgents tool (blue). Boxes: Q1/median/Q3; whiskers: 1.5\,IQR; circles: outliers; dots: individual trials ($n=20$).
  \label{fig:agentic-ai-metrics}}
\end{figure*}

All experiments in this section were conducted on the physical system, with real robot movement and part removal attempts. The Qwen 3.5 models (9B and 4B) run locally on the NVIDIA Jetson Thor, while GPT-4o-mini is accessed via API. We first evaluate a simple motion task: ``\emph{Move the robot 0.2m (forward/backward).}'' In the MCP condition, the agent discovers available capabilities through a ROS2 MCP server that exposes 39 generic services, and an additional targeted service \verb|/plan_to_pose_relative|. The agent must identify and invoke the correct service from this set. In the Tool condition, the same ROS2 service is wrapped as an explicit SmolAgents tool, removing the discovery step entirely. We run $n=20$ trials per model per mode (10 forward, 10 backward) across all three LLM backends, and compare the number of reasoning steps, input and output token counts, and wall-clock time to task completion. A trial is considered successful if the robot executes the commanded 0.2m displacement in the correct direction. Results are shown in Fig. \ref{fig:agentic-ai-metrics}.

With the MCP server, only Qwen 3.5 9B executed all 20 trials correctly. GPT-4o-mini failed once and Qwen 3.5 4B failed six times. Failures typically involve the model erroneously concluding task completion, sometimes even without attempting executions. MCP results also exhibit high variance, as the agent explores different subsets of the 39 available services across trials. Notably, the locally-deployed Qwen 3.5 9B matched or exceeded the cloud-based GPT-4o-mini in reliability, supporting the viability of fully offline operation. With the explicit tool interface, all three models completed all 20 trials correctly. All metrics improved substantially: input tokens dropped by an order of magnitude, and the GPT-4o-mini API cost fell from $\$0.24$ (MCP) to $\$0.04$ (Tool) across the 20 trials.

\begin{figure*}[!htb]
  \centering
  \includegraphics[width=\textwidth]{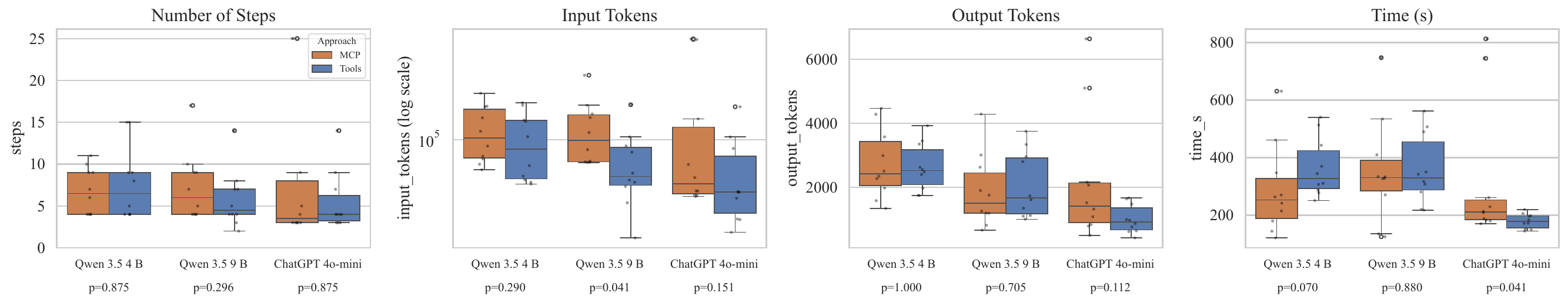}
  \caption{Metrics across three LLM back-ends for agentic execution of the complex screw-removal task (identify 5 of 68 remaining parts from the database, remove them, and update status) via MCP (orange) and an explicit SmolAgents tool (blue). Boxes: Q1/median/Q3; whiskers: 1.5\,IQR; circles: outliers; dots: individual trials ($n=10$ per approach). Although performance looks similar, MCP has a 43.3\% failure rate vs. 0\% for the tool-based approach.}
  \label{fig:agentic-ai-complex-metrics}
   \vspace{-12pt}
\end{figure*}

To evaluate complex reasoning, we prompt the agent: ``\emph{remove all the remaining parts from the top cover.}'' This requires the agent to query the JSON representation, identify all parts with the field \verb|is_removed| set to \emph{False}, execute a sequence of physical removal operations via the robot, and update each record to \emph{True}.  In all task instances, 5 of 68 parts remained to be removed. Both MCP and tool agents use the same inventory query tools; the only difference is how the physical removal action is exposed as a discoverable ROS2 service description in the MCP condition, or as an explicit SmolAgents tool in the Tool condition. Results are shown in Figure \ref{fig:agentic-ai-complex-metrics}.

Complex task reasoning performance differed sharply between MCP and Tool modes. Across 10 trials per model per mode (30 total per mode), the MCP approach produced 13 failures (43.3\%) while the Tool method produced none (0\%). Critically, 11 of the 13 MCP failures (36.7\% of all MCP trials) were false-success cases in which the agent marked all 5 screws as removed in the JSON inventory even though physical removal did not occur (e.g., 0/5 or 1/5 actually removed). False-success cases are a particular concern in high-voltage disassembly contexts. %; in deployment, we plan to gate inventory state updates on physical confirmation via nut-runner Hall effect current sensing, post-removal visual rechecking, and robot joint configuration monitoring. 
Because inventory management tools were identical across conditions, these failures demonstrate the advantage of explicit tool definitions over MCP-style discovery; while the MCP condition's 39-service search space partially conflates API-structure effects with action-space size, the complete elimination of failures under the tool interface suggests a structural benefit beyond search-space reduction alone.

Failure rates varied by model. Qwen 3.5 4B failed in 7/10 MCP trials, all false-success cases where 0/5 or 1/5 screws were physically removed despite 5/5 being recorded. Qwen 3.5 9B failed in 4/10 trials with the same pattern: 0/5 removed, 5/5 recorded. GPT-4o-mini had only 2/10 failures, and these were qualitatively different—all 5 screws were successfully removed, but 0/5 were tracked correctly in JSON, representing a tracking failure rather than a physical-removal failure. With explicit Tools, all three models achieved 5/5 removal with correct tracking across all 30 trials.

The total API cost for GPT-4o-mini across 10 MCP trials was approximately $\$0.21$ compared to $\$0.07$ for 10 Tool trials, reflecting the substantially higher token consumption of the MCP approach. When comparing execution time, successful Tool trials were generally faster because the agent batched all five removals into a single code block, whereas MCP agents tend to call services individually with re-discovery overhead between calls. However, MCP timing results are skewed toward shorter durations because failed trials where the agent skipped physical removal complete faster than successful ones. Overall, the tool-based architecture produced both higher task success rates and lower execution times for successful cases.

\section{DISCUSSION}\label{sec:discussion}
Consistent with \cite{qu2024robotic}, we observe that robots operate significantly slower than human workers (22~min vs.\ 17~min), although we run the robot well below its maximum speed. A success rate of $97\%$ would be sufficient for industrial deployment, pending validation that taught-in relative poses are robust to registration of the battery body. We anticipate improvements by using current sensing to detect engagement of the nut-runner and retrying if necessary, enhancing visual servoing and force control, and extending the agentic AI framework toward vision-language models to handle complex error conditions, such as missing or deteriorated screws.

For a 6+1~DoF system, planning still occasionally fails and does not always produce legible paths for human operators. Using corridor constraints produces predictable, straight end-effector motions, improving legibility at the cost of longer planning times. %Here, pre-computing and reusing motion plans for known battery types could improve both quality and speed. 

Our object detection models perform well on RGB data, but depth sensing struggles with reflective surfaces, such as the black sheet-metal battery cover and internal plastic components. Active infrared emission from the RealSense sensor leads to duplicate detections, which could be mitigated using recent depth-from-stereo machine learning approaches \cite{Abdelsalam2024Depth}.

Taught-in poses primarily fail due to end-effector slack, preventing 100\% success. Visual servoing improves accuracy, but exhibits different failure modes due to poor depth estimates, which produce control signals that either overshoot or undershoot, preventing convergence within the allowed iterations. In the future, we plan to investigate sensor fusion methods that allow off-line pose estimates that are comparable to taught-in ones. 

Agentic AI performance improves significantly when relevant services are provided as explicit ``tools'' to SmolAgents. Tools are not limited to robot operations---they can also define human-robot interactions, such as requesting a human to remove the battery cover or perform tasks beyond the robot's capabilities. MCP-based plans have higher failure rates, whereas no failures were observed when tools were explicitly defined. The key insight is that reliability comes from constraining the LLM's action space through well-defined tool interfaces with verified postconditions, not from the LLM's autonomy per se: the explicit tool wrapper ensures that inventory state updates occur only after confirmed physical execution. Selecting the right task abstractions is therefore a central design challenge for agentic AI programming of industrial robots.

\subsection{LIMITATIONS}
Several current limitations are targets for future work. Experiments are conducted on a single battery type, and success rates may not generalize to other battery designs or disassembly steps. All trials are single-shot without retry logic; reported success rates reflect first-attempt performance rather than practical throughput achievable with recovery behaviors. Depth sensing degrades on the reflective sheet-metal and plastic surfaces common in EV battery packs, causing occasional duplicate detections and poor depth estimates for visual servoing. Corridor-constrained path planning incurs substantial computational overhead (47~s average), limiting real-time responsiveness.

%One limitation is a deliberate design decision rather than an open problem: the MCP evaluation exposes 39 generic ROS2 services to characterize discovery-based interfaces under realistic conditions, which inflates the search space relative to a curated deployment. Our results support explicit tool interfaces as the preferred architecture for reliable agentic robot control; the MCP condition characterizes a failure mode rather than a deployment target.

\section{CONCLUSIONS}
We have presented a human-robot collaborative research platform for the disassembly of full-size electric vehicle batteries, focusing on the top-cover unscrewing task, which accounts for 10 of the 12 steps in disassembling a state-of-the-art 800V EV battery. The system is designed to be programmed using agentic AI, and we demonstrated two representative coding tasks that we evaluate in detail: executing basic robot motions and analyzing or altering task process memory. We observed that without structured tool interfaces, models may prematurely mark tasks as complete, causing discrepancies between the 
world model and reality; reliability comes from constraining the LLM's action space through well-defined tool interfaces with verified
postconditions, not from the LLM's autonomy per se.

Future work will extend the platform to perform full disassembly sequences with human-in-the-loop support, including handling parts that the robot cannot manipulate and assisting with tool changes. We also plan to incorporate a mobile manipulator to perform screw removal, tool exchange, and transport of small components, creating a flexible automation system that can be programmed via natural language. All design files and code are made available open-source to support further research and development.


%%%%%%%%%%%%%%%%%%%%%%%%%%%%%%%%%%%%%%%%%%%%%%%%%%%%%%%%%%%%%%%%%%%%%%%%%%%%%%%%

%%%%%%%%%%%%%%%%%%%%%%%%%%%%%%%%%%%%%%%%%%%%%%%%%%%%%%%%%%%%%%%%%%%%%%%%%%%%%%%%

\bibliographystyle{ieeetr}
\bibliography{battery}

%%%%%%%%%%%%%%%%%%%%%%%%%%%%%%%%%%%%%%%%%%%%%%%%%%%%%%%%%%%%%%%%%%%%%%%%%%%%%%%%

\section*{ACKNOWLEDGMENT}
%Sponsor information withheld for the purpose of double-blind review. 
This work has been supported by the Advanced Research Projects Agency-Energy (ARPA-E) CIRCULAR Program under cooperative agreement number DE-AR0001966.
Correll and Conway are with Realtime Manufacturing, Inc., a company that has an interest in the work presented here.

%%%%%%%%%%%%%%%%%%%%%%%%%%%%%%%%%%%%%%%%%%%%%%%%%%%%%%%%%%%%%%%%%%%%%%%%%%%%%%%%

\end{document}